%% file: main.tex
\documentclass[conference]{IEEEtran}
\IEEEoverridecommandlockouts

\usepackage{flushend}
\usepackage{balance}
\usepackage{booktabs}
\usepackage{multirow} 
\usepackage{diagbox}
  
\usepackage{amsmath,amsthm,amssymb,amsfonts}
\usepackage{algorithm}
\usepackage{algorithmic}
\usepackage{pifont}
\usepackage{bbding}
\usepackage{graphicx}
\usepackage{subfig}
\usepackage{pifont}
\usepackage{enumitem}
\usepackage[most]{tcolorbox}

\begin{document}

\title{
Multi-Agent System for Orchestrating Retrieval and Multimodal Reasoning in Cloud Environments
\thanks{\parbox[t]{\textwidth}{\IEEEauthorrefmark{3} Corresponding Author}}}

\author{
\IEEEauthorblockN{
Shuo Yang\IEEEauthorrefmark{1},
Xinran Zheng\IEEEauthorrefmark{2},
Jinfeng Xu\IEEEauthorrefmark{1}, 
Jinze Li\IEEEauthorrefmark{1}, 
Zewei Liu\IEEEauthorrefmark{1}, and
Edith C. H. Ngai\IEEEauthorrefmark{1}\IEEEauthorrefmark{3}}
\IEEEauthorblockA{\IEEEauthorrefmark{1}The University of Hong Kong, Hong Kong SAR, China}
\IEEEauthorblockA{\IEEEauthorrefmark{2}University College London, London, United Kingdom}}

\maketitle

\begin{abstract}
Intelligent reasoning services in cloud environments increasingly support knowledge-intensive tasks such as multimodal generation and fact-checking. While large language models have greatly improved individual reasoning capability, existing systems often rely on single-agent designs and fixed inference pipelines, which are brittle under unreliable retrieval and partial execution failures.
In this paper, we propose OR$^2$-MAS, a retrieval-augmented multi-agent framework designed to improve the robustness and efficiency of intelligent reasoning services through state-aware orchestration. OR$^2$-MAS coordinates retrieval and multi-agent reasoning through a global reasoning state, enabling adaptive routing, failure-aware redundancy, and dynamic control of reasoning depth. Experiments on multimodal fact-checking benchmarks show that OR$^2$-MAS achieves strong performance while remaining robust to retrieval and agent failures, highlighting the importance of system-level orchestration for cloud-based intelligent reasoning.
\end{abstract}


\maketitle

\input{Tex/1.Intro}
\input{Tex/2.Review}
\input{Tex/3.Method}
\input{Tex/4.Experiment}
\input{Tex/5.Conclusion}

\balance
\bibliographystyle{IEEEtran}
\bibliography{reference}

\end{document}

%% file: Tex/1.Intro.tex
\section{Introduction}
Recent years have witnessed a rapid proliferation of intelligent reasoning services~\cite{liang2022edge} deployed in cloud environments, driven by the growing demand for large-scale information analysis~\cite{nan2023large}, content moderation~\cite{markov2023holistic}, and decision support~\cite{soori2024ai}. Unlike traditional standalone AI models, these services are implemented as cloud-hosted reasoning pipelines that interact with heterogeneous, distributed, and network-bound data sources, while operating under constraints of elastic resource allocation, parallel execution, and low-latency inference. Consequently, intelligent reasoning is increasingly shifting from a model-centric optimization problem to a distributed system design challenge, where architectural choices, orchestration mechanisms, and retrieval pipelines critically determine system reliability and end-to-end effectiveness.

Advances in large language models (LLMs)~\cite{guo2024large, bai2025qwen2} have substantially improved the capability of individual agents to perform understanding and reasoning, achieving strong performance in planning~\cite{qiao2024agent,fei2024multimodal}, mathematical problem solving~\cite{didolkar2024metacognitive}, code generation~\cite{fakhoury2024llm}, and tool-augmented inference~\cite{chen2024advancing}. However, extending these capabilities to cloud environments introduces fundamental challenges~\cite{hu2025adaptive}. In cognitive computing tasks such as knowledge-intensive question answering~\cite{salemi2023symmetric}, fact-checking~\cite{choi2024fact}, and intelligence analysis~\cite{nan2023large}, the primary bottleneck is no longer computational capacity, but information incompleteness and reasoning uncertainty under dynamic execution conditions. Single-agent designs struggle to jointly handle evidence acquisition, cross-modal alignment, reasoning verification, and robustness in the presence of noise or partial failures~\cite{gao2025single}. In practice, transient retrieval failures or agent timeouts may propagate through the pipeline and invalidate the entire reasoning process~\cite{zhang2025agent}. Nevertheless, existing research focuses on model architectures~\cite{zhuge2024gptswarm} or prompt-level reasoning strategies~\cite{chen2024comm, xu2024mlevlm}, leaving system-level orchestration, fault tolerance, and adaptive control underexplored.

In this work, we formulate robust intelligent reasoning in cloud environments as a stateful and orchestrated system process. We propose OR$^2$-MAS, a \textbf{M}ulti-\textbf{A}gent \textbf{S}ystem that explicitly \textbf{O}rchestrates \textbf{R}etrieval and \textbf{R}easoning. The central design is to model the reasoning process through a global reasoning state that captures intermediate hypotheses, accumulated evidence, inter-agent consensus, uncertainty estimates, and agent execution status. 
This state serves as the basis for adaptive system control, enabling dynamic retrieval scheduling,
task routing among agents, failure-aware activation of backup agents, and adaptive reasoning termination. Rather than following a fixed inference pipeline, OR$^2$-MAS adjusts its reasoning depth and resource allocation according to real-time system feedback, thereby improving robustness and efficiency in cloud-based reasoning services.

We evaluate OR$^2$-MAS on a representative knowledge-intensive reasoning task, multimodal fact-checking, which requires dense web retrieval and cross-modal verification. The intrinsic dependence on accurate external knowledge and multi-step logical reasoning makes such workloads particularly challenging for intelligent reasoning systems. Experiments on benchmark datasets demonstrate that OR$^2$-MAS achieves strong performance while maintaining reliable multimodal reasoning under retrieval noise and agent failures. In summary, our contributions are as follows:
\begin{itemize}[left=0pt]
    \item We propose OR$^2$-MAS, a retrieval-augmented multi-agent system with adaptive routing and failure-aware control.
    \item We formulate intelligent reasoning as a stateful orchestration problem, highlighting the role of a global reasoning state in coordinating dynamic retrieval and multi-agent inference.
    \item We demonstrate through extensive experiments that OR$^2$-MAS achieves superior performance while maintaining robust multimodal reasoning under retrieval and agent failures.
\end{itemize}

%% file: Tex/2.Review.tex
\section{Related Work}

\subsection{Multi-Agent System}

Multi-agent systems (MAS)~\cite{li2024survey, chen2024internet} have been extensively studied as a paradigm for building complex intelligent systems through the interaction of multiple autonomous agents. Early work primarily focused on fixed agent teams with predefined workflows for downstream tasks~\cite{maldonado2024multi, chen2023autoagents, kim2024mdagents}. With the emergence of large-scale distributed computing, MAS designs have increasingly been explored to improve scalability, parallelism, and robustness~\cite{yang2024llm, qian2024scaling, wang2025megaagent}.

More recently, large language models (LLMs) have renewed interest in multi-agent systems for reasoning-intensive tasks~\cite{guo2024large, li2024survey}. These approaches typically exploit diversity among agents to improve reasoning quality through mechanisms such as self-consistency, voting, or role-based collaboration. However, most existing LLM-based multi-agent systems implicitly assume reliable execution environments and pay limited attention to system-level robustness~\cite{huangresilience, han2024llm}. In practical cloud deployments, agents may fail, time out, or exhibit unstable behavior due to resource contention or heterogeneous infrastructure~\cite{shetty2024building}, leading to cascading failures that degrade overall system reliability.

\subsection{Multimodal Reasoning and Fact-checking}

Multimodal reasoning~\cite{wang2024exploring, li2025perception} aims to enable models to jointly reason over heterogeneous modalities such as text, images, and videos. Advances in vision-language models~\cite{radford2021learning, li2019visualbert, dai2023instructblip} and multimodal large language models (MLLMs)~\cite{bai2025qwen2, bai2025qwen3vltechnicalreport, zhu2025internvl3} have substantially improved performance on tasks including visual question answering~\cite{xu2024mlevlm}, image--text matching~\cite{wu2024multimodal}, and multimodal generation~\cite{koh2023generating}. 

Many real-world multimodal reasoning tasks require access to up-to-date knowledge and multi-step evidence aggregation~\cite{yan2025position, wang2025multimodal}. Fact-checking~\cite{yao2023end, yang2025realfactbench} represents a representative and challenging application, where systems must assess cross-modal consistency and verify claims against external evidence. Early fact-checking methods primarily focused on textual analysis~\cite{Wadden2020, tang2024minicheck, choi2024automated}, while more recent work has incorporated images~\cite{papadopoulos2025red} and videos~\cite{cao2025video} to detect visually grounded misinformation. 

Recent LLM-based approaches further integrate external retrieval to support evidence grounding and explanation generation~\cite{ma2025local, qi2024sniffer, wu2025e2lvlm, yang2025rama}. Despite these advances, most existing systems rely on fixed inference pipelines, offering limited mechanisms to cope with retrieval noise or uncertainty during intermediate reasoning. These limitations motivate system-level approaches that emphasize adaptive retrieval, coordination, and robustness in multimodal fact-checking.

%% file: Tex/3.Method.tex
\section{Methodology}

\begin{figure*}
    \centering
    \includegraphics[width=1\linewidth]{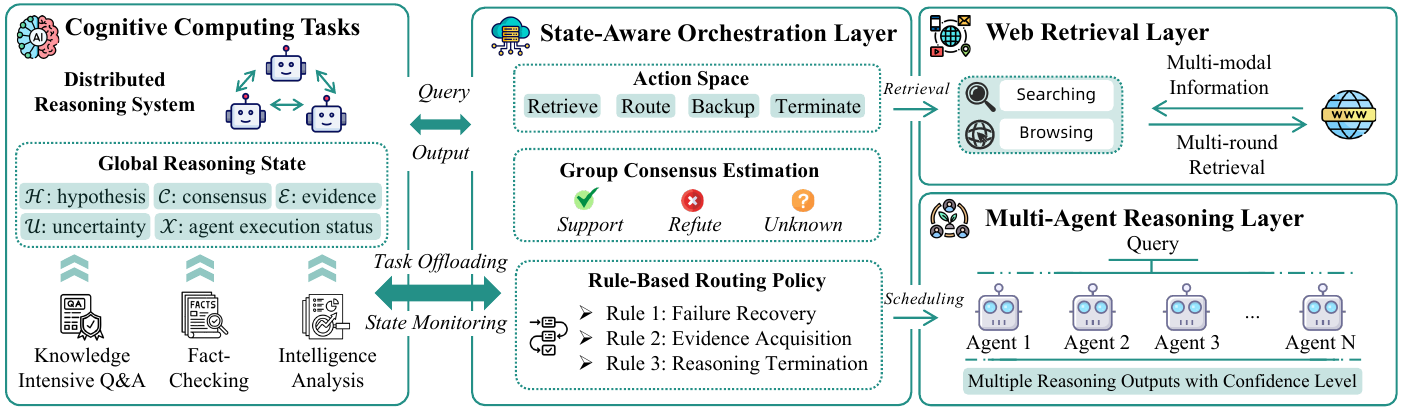}
    \caption{Overview of OR$^2$-MAS. OR$^2$-MAS is a state-aware multi-agent orchestration framework for cognitive cloud computing tasks. The system integrates three tightly coupled layers: a Web Retrieval Layer for multi-round evidence acquisition, a Multi-Agent Reasoning Layer composed of redundant and independent reasoning agents, and a State-Aware Orchestration Layer that coordinates retrieval, agent routing, failure recovery, and reasoning termination based on a global reasoning state.}
    \label{fig: overview}
\end{figure*}

OR$^2$-MAS is a multi-agent orchestration framework that treats retrieval and reasoning as dynamically coordinated, stateful services rather than a fixed inference pipeline. As illustrated in Fig.~\ref{fig: overview}, the framework is designed for cognitive computing tasks in cloud environments, such as knowledge-intensive question answering, fact-checking, and intelligence analysis, where reasoning must operate over distributed systems and unreliable execution environments.

The architecture of OR$^2$-MAS consists of three tightly integrated layers: (i) a Web Retrieval Layer that performs multi-round, on-demand acquisition of external evidence; (ii) a Multi-Agent Reasoning Layer composed of independent MLLM-based agents that generate diverse reasoning outcomes; and (iii) a State-Aware Orchestration Layer that monitors the global reasoning state and adaptively coordinates retrieval, agent routing, spawn backup, and termination. Crucially, the entire inference process is governed by a global reasoning state, which enables adaptive control over retrieval frequency, reasoning depth, and fault tolerance based on real-time uncertainty and execution feedback.

\subsection{System Model}

We formulate intelligent multimodal reasoning as an iterative, state-driven process, rather than a standalone model inference task. The system maintains a global reasoning state that summarizes the current hypothesis, accumulated evidence, inter-agent reasoning outcomes, and execution conditions. The following are the key concepts in the system model.

\subsubsection{\textbf{Multi-Agent Reasoning}}
Let $\mathcal{A} = \{a_1, a_2, \ldots, a_N\}$ denote a pool of reasoning agents. At reasoning round $t$, a subset $\mathcal{A}_t \subseteq \mathcal{A}$ is activated.
Each agent $a_i \in \mathcal{A}_t$ receives the same inputs and produces:
\begin{equation}
(y_i^{(t)}, c_i^{(t)}, h_i^{(t)}) \leftarrow a_i(q, \mathcal{E}_t, \mathcal{H}_t),
\end{equation}
where $y_i^{(t)} \in \mathcal{L}$ is the predicted verdict, $c_i^{(t)} \in [0,1]$ is a confidence score, and $h_i^{(t)}$ is a short local rationale. Agents do not share internal reasoning traces, ensuring logical independence and fault isolation.

\subsubsection{\textbf{Global Reasoning State}}

The reasoning process of OR$^2$-MAS is governed by a global state $\mathcal{S}_t$, which summarizes all information required for coordination:
\begin{equation}
\mathcal{S}_t = \{\mathcal{H}_t, \mathcal{E}_t, \mathcal{C}_t, \mathcal{U}_t, \mathcal{X}_t\}.
\end{equation}

Here, $\mathcal{H}_t$ denotes the current hypothesis or refined interpretation of the input query, $\mathcal{E}_t$ is the cumulative multimodal evidence set, $\mathcal{C}_t$ measures inter-agent consensus, $\mathcal{U}_t$ captures aggregated uncertainty, and $\mathcal{X}_t = \{ x_i^{(t)} |x_i^{(t)} \in \{\texttt{success}, \texttt{timeout}, \texttt{failure}, \texttt{pending}\} \}$ records execution states of agents.

\subsubsection{\textbf{Action Space}}

At each reasoning round, the system selects an action $a_t$ from a discrete action space:
\begin{equation}
\Omega = \{\texttt{retrieve},\ \texttt{backup},\ \texttt{route},\ \texttt{terminate}\}.
\end{equation}

Actions affect the evolution of the global state, leading to the next state $\mathcal{S}_{t+1}$. The reasoning process terminates either when a termination action is selected or when a maximum number of rounds is reached.

\subsection{Web Retrieval Layer}

The retrieval layer serves as the system’s external perception module, responsible for acquiring up-to-date and relevant multimodal evidence from open web sources or domain-specific knowledge repositories (e.g., vector databases or document stores). It includes two well-defined tools:

\begin{itemize}[left=0pt]
    \item Searching: The tool searches for relevant information on the web based on a set of search queries, and then returns the top-$k$ results, each including the title, URL, and summary.
    \item Browsing: For each retrieved URL in browsing requests, this tool extracts content not encountered in the dialogue history. A short-term memory buffer is maintained for each query to avoid repeated browsing and enhance evidence diversity.
\end{itemize}

At round $t$, search and browse tools can be invoked multiple times throughout the retrieval process:
\begin{equation}
\mathcal{E}_t = \mathcal{E}_{t-1} \cup \textsc{Retrieve}(q, \mathcal{H}_{t-1}, k),
\end{equation}
where $\mathcal{H}_{t-1}$ includes refinement of input query $q$, including searching and browsing requests, $k$ controls the per-round evidence budget, and only the top-$k$ results are retained to limit context growth in downstream reasoning agents.

\subsection{Multi-Agent Reasoning Layer}

The reasoning layer comprises a pool of functionally equivalent but logically isolated MLLM agents. Each agent operates as an independent execution unit and receives the same inputs. Redundancy is introduced for fault tolerance rather than functional decomposition, and agents do not share internal reasoning traces, ensuring that failures or degradations of individual agents do not compromise the entire system.

We formalize each agent as $ a_i = \{x_i, m_i\}$, where $x_i $ denotes the status of agents, and $m_i$ denotes the used LLM backbone (e.g., Qwen2.5-VL~\cite{bai2025qwen2}, InternVL3~\cite{zhu2025internvl3}). In this layer, differentiated deployment of models with different types and sizes can be considered to provide more efficient choices for reasoning difficulty. By executing multiple agents independently and aggregating their outcomes, the system introduces redundancy at the execution level, which improves resilience to stochastic model behavior, partial execution failures, and transient performance degradation. This design reduces reliance on any single agent instance and supports stable decision-making under dynamic execution environments.

\subsection{State-Aware Orchestration Layer}

The orchestration layer acts as a central controller,
dynamically coordinating retrieval and multi-agent reasoning based on the current global state $\mathcal{S}_t$. Rather than following a fixed pipeline, the orchestrator adapts both execution paths and reasoning depth in response to observed consensus, uncertainty, and failures. 

\subsubsection{\textbf{Group Consensus Estimation}}

Let $\mathcal{L}$ denote the predefined label space for the fact-checking task
(e.g., $\mathcal{L}=\{\texttt{support},\texttt{refute},\texttt{unknown}\}$).
At reasoning round $t$, each active agent $a_i \in \mathcal{A}_t$
produces a predicted label $y_i^{(t)} \in \mathcal{L}$
with an associated confidence score $c_i^{(t)} \in [0,1]$.

We quantify inter-agent consensus as the maximum fraction of agents
agreeing on the same label:
\begin{equation}
\mathcal{C}_t
= \max_{\ell \in \mathcal{L}}
\frac{1}{|\mathcal{A}_t|}
\sum_{a_i \in \mathcal{A}_t}
\mathbb{I}\!\left[y_i^{(t)} = \ell\right],
\end{equation}
where $\mathbb{I}[\cdot]$ is the indicator function.
$\mathcal{C}_t \in [0,1]$ measures the degree of agreement among agents, with higher values indicating stronger consensus.

To complement consensus with confidence-aware information, we define an aggregated uncertainty score that focuses on the confidence of the majority-supported prediction.
Let
\begin{equation}
\ell_t^{*}
= \arg\max_{\ell \in \mathcal{L}}
\sum_{a_i \in \mathcal{A}_t}
\mathbb{I}\!\left[y_i^{(t)} = \ell\right]
\end{equation}
denote the majority label at round $t$.
We then define aggregated uncertainty as:
\begin{equation}
\mathcal{U}_t
= 1 -
\frac{1}{|\mathcal{A}_t|}
\sum_{a_i \in \mathcal{A}_t}
\mathbb{I}\!\left[y_i^{(t)} = \ell_t^{*}\right]
\cdot c_i^{(t)}.
\end{equation}

Here, $\mathcal{U}_t$ decreases when a larger fraction of agents
not only agree on the same label but also assign it high confidence.
In contrast, high disagreement or low-confidence agreement
results in higher uncertainty.

Together, $\mathcal{C}_t$ and $\mathcal{U}_t$
provide complementary system-level signals:
$\mathcal{C}_t$ captures the structural agreement among agents,
while $\mathcal{U}_t$ reflects the reliability of that agreement.
These signals are subsequently used by the orchestration layer
to control retrieval, redundancy, and termination decisions.

\subsubsection{\textbf{Rule-Based Routing Policy}}

We implement orchestration using a rule-based policy that maps the current state $\mathcal{S}_t$ to an action $a_t$. The routing policy is designed to prioritize system robustness and reasoning efficiency under dynamic execution conditions.

Rule 1: Failure Recovery. To ensure system availability, failure recovery is always prioritized. Let $\mathcal{X}_t$ denote agent execution states at round $t$. If the proportion of failed or timed-out agents exceeds a threshold $\beta$,
\begin{equation}
\label{eq:failure_ratio}
\frac{1}{|\mathcal{A}_t|}
\sum_{a_i \in \mathcal{A}_t}
\mathbb{I}\!\left[x_i^{(t)} \in \{\texttt{failure}, \texttt{timeout}\}\right] > \beta,
\end{equation}
the orchestrator first triggers \texttt{backup} to activate redundant agents, and subsequently applies \texttt{route} to redirect pending subtasks away from failed executions. This rule prevents cascading failures caused by isolated node unavailability.

Rule 2: Evidence Acquisition. When the reasoning process exhibits instability, additional evidence is prioritized as a lightweight corrective action. Specifically, if either inter-agent consensus is low or aggregated uncertainty is high,
\begin{equation}
\label{eq:evidence_insufficiency}
\mathcal{C}_t < \tau_c
\;\vee\;
\mathcal{U}_t > \tau_u,
\end{equation}
the orchestrator triggers \texttt{retrieve} to acquire supplementary evidence guided by the current hypothesis $\mathcal{H}_t$. Then the orchestrator may additionally apply \texttt{route} to assign the task to alternative agents,
thereby diversifying reasoning paths rather than relying on deeper execution of the same agents.

Rule 3: Reasoning Termination. Reasoning depth is controlled implicitly through repeated application of rule 2. When sufficient agreement and confidence are achieved,
\begin{equation}
\label{eq:termination_condition}
\mathcal{C}_t \ge \tau_c
\;\wedge\;
\mathcal{U}_t \le \tau_u,
\end{equation}
the orchestrator selects \texttt{terminate} and outputs the aggregated verdict. This early termination criterion enables simple or well-supported claims to converge in a small number of rounds, while reserving additional retrieval and redundancy for ambiguous or conflicting cases.

%% file: Tex/4.Experiment.tex
\section{Experiment }

\subsection{Experimental Setups}

\subsubsection{\textbf{Dataset}}
We evaluate our method on the NewsCLIPpings~\cite{luo2021newsclippings} dataset, which contains realistic image--text pairs reflecting common visual misinformation. Following prior work~\cite{qi2024sniffer}, we report overall and class-wise accuracy.

\subsubsection{\textbf{Baselines}}

We compare OR$^2$-MAS with representative methods from three paradigms: from-scratch models~\cite{zhou2020similarity}, pretrained multimodal approaches~\cite{zhang2023detecting}, and evidence-augmented LLM-based systems~\cite{qi2024sniffer, wu2025e2lvlm}. Specifically, SAFE~\cite{zhou2020similarity} measures image--text relevance via sentence-level similarity. Neu-Sym~\cite{zhang2023detecting} adopts a neural--symbolic framework that decomposes claims into fact queries evaluated by CLIP~\cite{radford2021learning}. SNIFFER~\cite{qi2024sniffer} leverages external evidence generated by GPT-4 to fine-tune an InstructBLIP model~\cite{dai2023instructblip}. E2LVLM~\cite{wu2025e2lvlm} retrieves web evidence via reverse image search and jointly reasons over visual and textual inputs using an MLLM.

\subsubsection{\textbf{Implementation Details}}

OR$^2$-MAS is implemented as a modular framework, where retrieval, reasoning, and orchestration are decoupled as independent services. The retrieval layer is instantiated using DeepResearcher-7B~\cite{zheng2025deepresearcher}, which supports iterative web search and browsing. By default, the reasoning layer consists of three concurrently executed agents, each instantiated from a $\sim$70B-scale multimodal language model, including Qwen2.5-VL~\cite{bai2025qwen2}, Qwen3-VL~\cite{bai2025qwen3vltechnicalreport}, and InternVL3~\cite{zhu2025internvl3}. To enhance robustness against execution failures, we additionally deploy three backup agents instantiated from smaller $\sim$30B-scale models, which are activated on demand. The orchestration is driven by the global reasoning state, with conservative thresholds to prioritize robustness. Specifically, the failure threshold $\beta$ is set to 0, ensuring immediate recovery upon any agent failure, while the consensus and uncertainty thresholds are set to $\tau_c=0.6$ and $\tau_u=0.2$, respectively. 

\subsection{Performance Comparison}

\input{Tab/main_results}

We first evaluate the overall effectiveness of OR$^2$-MAS on the NewsCLIPpings dataset. As reported in Table~\ref{tab: news}, OR$^2$-MAS achieves the highest overall accuracy among all compared methods. Early approaches that do not leverage pretrained multimodal representations or external evidence, such as SAFE, exhibit substantially lower performance, underscoring the challenges of multimodal misinformation detection under limited semantic supervision. Neu-Sym Detector benefits from pretrained vision-language models and achieves a significant improvement over SAFE. Methods that incorporate external evidence retrieval further boost performance. Notably, compared with E2LVLM, which relies on a single-turn agent workflow, OR$^2$-MAS attains higher accuracy, demonstrating the effectiveness of multi-round, state-aware orchestration over one-shot retrieval and reasoning.

\subsection{Ablation Study}

\input{Tab/ablation}

We conduct an ablation study to analyze the contribution of key components in OR$^2$-MAS. Removing evidence retrieval (OR$^2$-MAS w/o R) leads to a significant performance drop, confirming the importance of retrieval augmentation. OR$^2$-MAS w/o O adopts a one-time retrieval and reasoning strategy with simple majority voting. Disabling state-aware orchestration in this manner results in degradation, demonstrating that orchestration plays a critical role in adaptive coordination.

\subsection{Robustness Analysis}

To evaluate robustness under realistic deployment conditions, we simulate failures and degradations commonly encountered in distributed inference settings.

\begin{figure}[!t]
\centering
\subfloat[Retrieval Failure]{
\includegraphics[height=2.7cm]{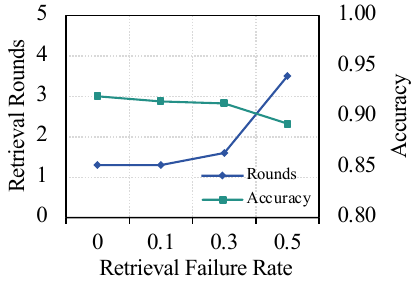}
\label{fig: retrieval}} 
\subfloat[Agent Failure]{
\includegraphics[height=2.7cm]{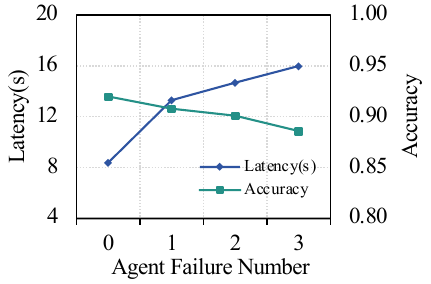}
\label{fig: agent}}
\caption{Robustness evaluation under common failures.}
\label{fig:abalation}
\end{figure}

\subsubsection{\textbf{Retrieval Failure}}
We simulate retrieval degradation by randomly dropping retrieved evidence or replacing a portion of retrieved documents with irrelevant web content, with the noise ratio varied from 0.1 to 0.5. As shown in Fig.~\ref{fig: retrieval}, OR$^2$-MAS maintains stable performance under increasing noise. When the failure rate reaches 0.5, the average number of retrieval rounds largely increases, indicating that the orchestrator adaptively triggers additional retrieval to compensate for degraded evidence quality.

\subsubsection{\textbf{Agent Failure}}

We further evaluate robustness under agent failures by randomly forcing a subset of reasoning agents to time out or return invalid outputs (see Fig.~\ref{fig: agent}). Upon detecting failures, OR$^2$-MAS activates backup agents instantiated from smaller $\sim$30B-scale models to preserve system availability. While replacing larger models with smaller ones may introduce a mild degradation in reasoning capacity, the overall accuracy remains largely stable. This result indicates that failure-aware redundancy and dynamic rerouting effectively mitigate the impact of agent failures, allowing the system to sustain reliable reasoning performance despite heterogeneous agent capabilities and execution disruptions.

%% file: Tab/main_results.tex
\begin{table}[t]
\setlength{\tabcolsep}{4.5pt}
    \centering
    \caption{Comparison results on benchmark dataset. The best results are highlighted in bold.}
    \label{tab: news}
\begin{tabular}{ccccccc}
\toprule
\textbf{Methods} & \textbf{Retrieval} & \textbf{VLM} & \textbf{Multi-Agent} &\textbf{All} & \textbf{Fake}  & \textbf{Real}  \\ \midrule
SAFE~\cite{zhou2020similarity}            & \ding{55}                  & \ding{55}  & \ding{55} & 0.528    & 0.548 & 0.520  \\
Neu-Sym~\cite{zhang2023detecting}        & \ding{55}                 & \ding{51}  & \ding{55}  & 0.682    &  $-$     &   $-$    \\
SNIFFER~\cite{qi2024sniffer}         & \ding{51}                  & \ding{51}   & \ding{55} & 0.884    & 0.869 & 0.918 \\
E2LVLM~\cite{wu2025e2lvlm}          & \ding{51}                   & \ding{51}  & \ding{55}  & 0.899    & 0.903 & 0.894 \\ \midrule
OR$^2$-MAS          & \ding{51}                   & \ding{51}  & \ding{51}  & \textbf{0.920}     &\textbf{0.914} & \textbf{0.926} \\ \bottomrule
\end{tabular}
\end{table}

%% file: Tab/ablation.tex
\begin{table}[t]
    \setlength{\tabcolsep}{10pt}
    \centering
    \caption{Ablation study. R: Retriever. O: Orchestration.}
    \label{tab:ablation_study}
    \begin{tabular}{cccc}
         \toprule
      \textbf{Methods}       & \textbf{All}   & \textbf{Fake}  & \textbf{Real}  \\ \midrule
OR$^2$-MAS w/o R & 0.674 & 0.529 & 0.818 \\ \midrule
OR$^2$-MAS w/o O & 0.898 & 0.891 & 0.905 \\ \midrule
OR$^2$-MAS       & 0.920  & 0.914 & 0.926 \\ 
         \bottomrule
    \end{tabular}
\end{table}

%% file: Tex/5.Conclusion.tex
\section{Conclusion}

In this paper, we study intelligent multimodal reasoning from a system-level perspective and argue that robust reasoning in cloud environments is a stateful and orchestrated process rather than a single model inference task. To this end, we propose OR$^2$-MAS, a retrieval-augmented multi-agent framework that explicitly coordinates web retrieval and reasoning through a global reasoning state. By enabling adaptive routing, failure-aware redundancy, and dynamic control of reasoning depth, OR$^2$-MAS effectively addresses the instability and uncertainty inherent in cloud-based reasoning services. We evaluate OR$^2$-MAS on multimodal fact-checking, a representative knowledge-intensive reasoning task. Experimental results demonstrate that OR$^2$-MAS not only achieves strong performance but also maintains robust reasoning under retrieval noise and agent failures, validating the effectiveness of system-level orchestration. We believe this work highlights the importance of orchestration as a first-class design principle for scalable and reliable intelligent reasoning systems and opens avenues for applying similar orchestration-driven approaches to broader cognitive cloud computing tasks.